\documentclass{article}
\PassOptionsToPackage{numbers,sort&compress}{natbib}
\usepackage[preprint]{neurips_2026}
\makeatletter
\renewcommand{\@noticestring}{}
\renewcommand{\@notice}{}
\makeatother
\usepackage[utf8]{inputenc}
\usepackage[T1]{fontenc}
\usepackage{float}
\usepackage[hidelinks]{hyperref}
\usepackage{booktabs}
\usepackage{graphicx}
\usepackage{xcolor}
\usepackage{tikz}
\usetikzlibrary{positioning,arrows.meta,calc,fit,shapes,backgrounds}
\usepackage{array}
\usepackage{amsmath}
\usepackage{multirow}
\usepackage{enumitem}
\usepackage{caption}
\usepackage{xspace}
\setlist{nosep,leftmargin=1.4em}
\newcolumntype{L}[1]{>{\raggedright\arraybackslash}p{#1}}
\newcommand{\rhead}[1]{\textbf{#1}\enspace}
\newcommand{\qualdir}{figs/}
\newcommand{\ms}{\,m/s\xspace}

\graphicspath{{figs/}}

\newcommand{\sysname}{GuardLTF\xspace}
\newcommand{\decname}{road-boundary-aware decoder\xspace}
\newcommand{\Decname}{Road-boundary-aware decoder\xspace}
\newcommand{\decshort}{boundary-aware decoder\xspace}
\newcommand{\Decshort}{Boundary-aware decoder\xspace}
\newcommand{\declossname}{road-boundary loss\xspace}
\newcommand{\dlname}{risk-aware decision layer\xspace}
\newcommand{\Dlname}{Risk-aware decision layer\xspace}
\newcommand{\base}{\mbox{GuardLTF (safeguards)}\xspace}
\newcommand{\early}{\mbox{GuardLTF (lite)}\xspace}
\newcommand{\stD}{\mbox{GuardLTF (+decoder)}\xspace}
\newcommand{\stL}{\mbox{GuardLTF (+lead)}\xspace}
\newcommand{\stC}{\mbox{GuardLTF (+creep)}\xspace}

\newcommand{\cfgtgt}{+\,Targeted checks\xspace}
\newcommand{\cfgrel}{+\,Release and launch\xspace}
\newcommand{\cfgbrk}{+\,Braking cap and pace\xspace}
\newcommand{\cfgcrp}{+\,Free-space creep\xspace}
\newcommand{\cfgrol}{+\,Launch for rolling starts\xspace}
\newcommand{\cfgfar}{+\,Creep behind a far lead\xspace}

\newcommand{\papertitle}{Guarded Latent TransFuser with a Risk-Aware Decision Layer: AlpaSim E2E Closed Loop Challenge Technical Report}
\hypersetup{pdftitle={\papertitle}, pdfauthor={Jiahao Shu, Zhiqing Hong, Baoshen Guo}}

\title{\papertitle}
\author{%
  \makebox[\dimexpr\linewidth-12pt\relax][c]{\begin{tabular}[t]{@{}c@{\hspace{2.4em}}c@{\hspace{2.4em}}c@{}}
    \textbf{Jiahao Shu} & \textbf{Zhiqing Hong} & \textbf{Baoshen Guo}$^{\dagger}$\\
    \normalfont Carnegie Mellon University & \normalfont The Hong Kong University of Science & \normalfont Singapore-MIT Alliance for\\
     & \normalfont and Technology (Guangzhou) & \normalfont Research and Technology
  \end{tabular}}%
}

\begin{document}
\maketitle
\begingroup\renewcommand{\thefootnote}{}\footnotetext{$^{\dagger}$Corresponding author: \texttt{baoshen.guo@smart.mit.edu}}\endgroup

\begin{abstract}
In this technical report, we present \textit{\sysname}, a camera-only Guarded Latent TransFuser planner with a risk-aware decision layer, with which our team, SymPhi, took first place in the nuPlan track of the AlpaSim E2E Closed Loop Challenge 2026.
Through an analysis of closed-loop failures, we highlight two key considerations for this track:
(i) imperfect camera-only perception, particularly of road boundaries; and
(ii) the risk-sensitive nature of closed-loop planning, where a single at-fault collision or off-road exit results in a zero score for the entire scene.
To address these challenges, we build \sysname on the frozen Latent TransFuser (LTF) with two components:
(i) a \emph{\decname}, fine-tuned from LTF's trajectory decoder to keep planned trajectories within the drivable area, with the mapped drivable area used as a training label only; and
(ii) a \emph{\dlname} that uses lead-vehicle motion and free-space estimates from a near-field camera detector to guide planning decisions.
In paired closed-loop evaluation on a validation set of 3,625 scenes, \sysname reduces the incident scenes (off-road exit or at-fault collision) of the LTF planner from 103 to 45 and raises the mean scene score from 0.936 to 0.957 (95\,\% confidence interval of the paired difference $[+0.010, +0.028]$).
On the official test set, \sysname reaches a mean scene score of 0.9763 and drives 2.28\,km per incident, compared with 0.9205 and 0.83\,km for the unchanged LTF planner.
Code is available at \url{https://github.com/JiahaoShu/GuardLTF}.
\end{abstract}

\section{Introduction}
\label{sec:intro}
End-to-end planners map camera frames to a trajectory with a single network trained by imitation~\citep{chitta2023transfuser,hu2023uniad,chen2024e2esurvey}.
The nuPlan track of the AlpaSim E2E Closed Loop Challenge~\citep{alpasimchallenge} evaluates such a camera-only policy in closed loop, in a neural-rendering simulator, on 5\,s scenes reconstructed from nuPlan drives~\citep{caesar2021nuplan}.
Its score is risk-sensitive: an at-fault collision, an off-road exit or an exit from a corridor around the human trajectory sets the score of the scene to zero, and otherwise the score is the progress relative to the human driver.
Failed scenes therefore weigh heavily in the score.

Through an analysis of closed-loop failures (Appendix~\ref{app:diagnosis}), we identify two limitations of the Latent TransFuser (LTF) planner~\citep{chitta2023transfuser,py123dgarage}:
\textit{(i) Camera-only perception is imperfect.}
The planner's bird's-eye-view (BEV) map does not accurately capture road boundaries.
Most zero scores are off-road exits in which the planned path crosses the road edge by a few decimetres, at places where the BEV map still predicts road; a check on the predicted map therefore repairs only a minority of them.
\textit{(ii) The planner is not sufficiently responsive to surrounding vehicles and environmental cues.}
For example, it continues driving where the human driver stops behind a vehicle, and when starting from rest it often pulls away before the human driver does.
We therefore design \sysname, which combines a \decname with a \dlname around the frozen LTF backbone (Fig.~\ref{fig:overview}).
Our main contributions are as follows:
\begin{itemize}
\item To address imperfect camera-only perception, we design a \emph{\decname}, fine-tuned from the planner's trajectory decoder with a hinge loss on the distance between the planned vehicle footprint and the drivable-area boundary. The mapped drivable area is used only as a training label, so no map is needed at run time.
\item To address the insufficient response to surrounding vehicles and environmental cues, we design a \emph{\dlname}. It applies gated switching between the two trajectory decoders, and adjusts the selected trajectory using lead-vehicle motion and the free-space estimate of a near-field camera detector.
\item In paired closed-loop evaluation on 3,625 validation scenes, \sysname more than halves the incident scenes of the LTF planner, and on the official test set it raises the mean scene score from 0.9205 to 0.9763. We report an ablation of every component. Our team took first place in the nuPlan track of the challenge.
\end{itemize}

\rhead{Related work.}
TransFuser~\citep{chitta2023transfuser} fuses sensor features with transformers and predicts waypoints together with auxiliary BEV outputs; its camera-only variant LTF is the planner used here, and UniAD~\citep{hu2023uniad} and VAD~\citep{jiang2023vad} train perception, prediction and planning jointly.
Policies trained by behaviour cloning degrade when they drive on their own plans~\citep{codevilla2019limitations}, and on nuPlan a rule-based planner remains competitive with learned ones in closed loop~\citep{dauner2023pdm}.
Planning safety has been addressed with collision and drivable-area losses in ChauffeurNet~\citep{bansal2019chauffeurnet} and VAD~\citep{jiang2023vad}, with learned trajectory scoring in Hydra-MDP~\citep{li2024hydramdp} and GTRS~\citep{li2025gtrs}, and with explicit safety layers around learned policies~\citep{shao2022interfuser,alshiekh2018shielding}.
Our decoder follows the drivable-area losses, and its training adds the policy's own rollouts in the spirit of DAgger~\citep{ross2011dagger}; in contrast to these methods, the planner stays frozen, and the decoder is applied selectively at run time by an explicit decision layer.

\section{Method summary}
\label{sec:method}
\begin{figure}[h]
\centering
\input{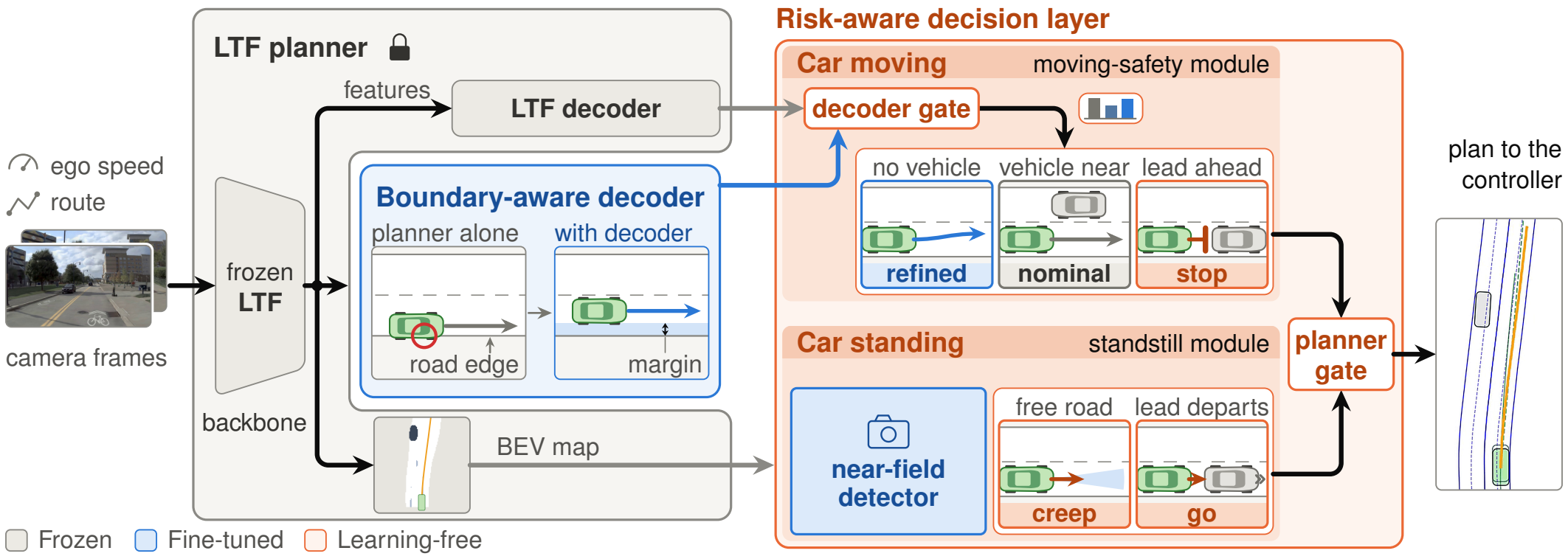}
\caption{Overview of \sysname. The frozen LTF planner produces the \emph{nominal} plan (grey) and a BEV map; the \decshort produces the \emph{refined} plan (blue) from the same features. While the car moves, the decoder gate selects one of the two plans and the moving-safety module may stop it behind a lead vehicle; while the car stands, the standstill module decides between holding, creeping and departing. The planner gate passes on the output of the modules that match the car's state.}
\label{fig:overview}
\end{figure}

\sysname (Fig.~\ref{fig:overview}) keeps the public LTF planner frozen and adds a second trajectory decoder and a decision layer.
At each replan the frozen planner produces a 6.4\,s trajectory, the \emph{nominal} plan, and a BEV map with road and vehicle predictions.
From the same backbone features, the \decshort produces a second trajectory, the \emph{refined} plan, which keeps a margin to the road edge.
The decision layer selects one of the two plans, using the BEV map and, while the car stands, a near-field camera detector, and passes it to the controller.

\section{Policy description}
\label{sec:policy}
\rhead{Inputs and outputs.}
A scene consists of a \emph{hand-over}, a replayed segment of the recorded drive, followed by 5\,s under policy control with a replan every $T = 0.5$\,s.
At each replan the policy receives two front-camera frames 1\,s apart, the ego speed and route points, and returns a trajectory that a controller tracks.

\rhead{\Decname.}
\label{sec:decoder}
The \decshort is initialised as a copy of the LTF trajectory decoder and fine-tuned while the backbone and all other weights stay frozen, so both decoders read the same backbone features.
The loss is
\begin{equation}
\mathcal{L} = \lambda_{\mathrm{s}}\,\mathcal{L}_{\mathrm{state}} + \lambda_{\ell}\,\mathcal{L}_{\mathrm{len}} + \lambda_{\mathrm{h}}\,\mathcal{L}_{\mathrm{h}} + \lambda_{\mathrm{a}}\,\mathcal{L}_{\mathrm{a}},
\qquad
\mathcal{L}_{\mathrm{h}} = \frac{\sum_k w_k \max(0,\; m - d_k)}{\sum_k w_k},
\label{eq:loss}
\end{equation}
where $d_k$ is the signed distance of the planned vehicle footprint to the drivable-area boundary at plan step $k$, $m = 0.15$\,m is the margin and $w_k$ down-weights steps beyond 3\,s.
The hinge $\mathcal{L}_{\mathrm{h}}$ follows the drivable-area losses of ChauffeurNet and VAD~\citep{bansal2019chauffeurnet,jiang2023vad}; $\mathcal{L}_{\mathrm{a}}$ is a hinge of the same kind that keeps the planned footprint $m_{\mathrm{a}} = 0.5$\,m away from other road users.
The two distillation terms keep the end state ($\mathcal{L}_{\mathrm{state}}$) and the arc length ($\mathcal{L}_{\mathrm{len}}$) of the plan close to those of the LTF decoder.
The drivable area of the nuPlan maps enters only through $d_k$ during training.
The decoder is trained on recorded nuPlan drives and on closed-loop rollouts of our own system: a decoder trained on recorded drives alone sees only states on the human trajectory, and the rollouts add the states that the policy itself reaches~\citep{ross2011dagger}.
The ablation also uses a \emph{preliminary decoder}, trained on recorded drives only and without $\mathcal{L}_{\mathrm{a}}$; where the two are compared, we call the decoder of \sysname the \emph{final decoder}.

\rhead{\Dlname.}
\label{sec:decision}
The decision layer consists of three modules; a planner gate passes on the output of those that match the car's state, moving or standing (Fig.~\ref{fig:overview}). A \emph{lead vehicle} is a vehicle that the BEV map predicts in the lane ahead of the car.
\begin{itemize}[leftmargin=1.1em, itemsep=1pt, topsep=1pt, parsep=0pt, partopsep=0pt]
\item \textbf{Decoder gate.} While the car moves, the gate selects the refined plan by default. It falls back to the nominal plan on the first replan of a scene, for a short time after the BEV map shows a vehicle close to the planned path (\emph{proximity gate}), and at high speed (\emph{speed gate}). The gate limits the collisions that a lateral shift next to another vehicle can cause.
\item \textbf{Moving-safety module.} The module checks the selected plan against the BEV map. If the car closes in on a lead vehicle with a short time to collision, or a road user is predicted inside the planned footprint, the plan is shortened so that the car stops ahead of it. A plan that leaves the predicted road is replaced by the closest plan that stays on it. Otherwise the module only re-times the plan: a \emph{braking cap} limits the travel distance when the hand-over shows hard braking behind a slow lead, a \emph{launch for rolling starts} lets a slowly rolling car that was accelerating at the hand-over continue to accelerate, and a \emph{pace gain} of 3\,\% applies on a free, straight road.
\item \textbf{Standstill module.} A car that stands at the end of the hand-over is held; we call such a scene a \emph{held start}. The car is released when the planner's own plan accelerates and the lead vehicle has moved away, and behind a near lead it then follows the lead at a fixed time headway (\emph{release and launch}). Where the BEV map shows no lead, or only a far one, the car creeps forward by at most $L_{\mathrm{c}} = 3.5$\,m as long as the near-field detector reports a free road for at least $d_{\mathrm{f}} = 8$\,m at every replan (\emph{free-space creep}). Because $L_{\mathrm{c}} < d_{\mathrm{f}}$, the creep never reaches road that the detector has not confirmed.
\end{itemize}
Four of these behaviours are simple checks on the planner's own outputs and need neither the second decoder nor the detector: the hold of a standing car, its release once the lead vehicle has moved away, the stop behind a lead vehicle or a crossing road user, and the guard that keeps the plan on the predicted road. We call them the \emph{basic safeguards}.

The \emph{near-field detector} is a SegFormer MiT-B0 encoder~\citep{xie2021segformer}, initialised from a public checkpoint fine-tuned on Cityscapes~\citep{cordts2016cityscapes}, with a head of ordered outputs for the gap ahead of the bumper. It is trained on front-camera frames rendered in our closed-loop rollouts, with the actor boxes of the simulator as labels, and runs only while the car is held (Table~\ref{tab:detector}).

\rhead{Variants.}
LTF denotes the public checkpoint without any change; every other configuration is a variant of \sysname and is named by what it contains.
\base is the frozen LTF planner with only the basic safeguards of the decision layer and no second decoder.
\stD, \stL and \stC add components to it in cumulative stages (Table~\ref{tab:main}), and \sysname is the full system.
\early is a lighter version with only the hold, the release and the road guard; it is the entry with which our team took first place (Table~\ref{tab:board}).

\section{Results}
\label{sec:results}
\rhead{Evaluation protocol.}
Local results are paired closed-loop runs in the official simulator, controller and scorer~\citep{alpasimcode}, with one rollout per scene.
We use 11,146 scenes of the official scene packages: 7,521 \emph{development scenes} and a \emph{validation set} of 3,625 scenes from 21 recording sessions, which share no scene (Appendix~\ref{app:sets}).
The \emph{test set} is the hidden set of 1,000 scenes behind the official leaderboard.
An \emph{incident scene} is a scene with an at-fault collision or an off-road exit.
Confidence intervals are bootstrap intervals over recording sessions.
Table~\ref{tab:main} starts from the public LTF checkpoint and \base and adds the components in four cumulative stages; Table~\ref{tab:steps} has every step.

\begin{table}[!htb]
\centering
\caption{Ablation on the validation set (3,625 scenes). The stages below \base are cumulative. Incident scenes: scenes with an at-fault collision or an off-road exit. Distance: km driven per incident scene.}
\label{tab:main}
\small
\setlength{\tabcolsep}{4.2pt}
\begin{tabular}{@{}llrrrr@{}}
\toprule
Configuration & Added components & \begin{tabular}[b]{@{}r@{}}Incident\\scenes\end{tabular} & \begin{tabular}[b]{@{}r@{}}At-fault\\collisions\end{tabular} & \begin{tabular}[b]{@{}r@{}}Mean\\score\end{tabular} & \begin{tabular}[b]{@{}r@{}}Distance\\(km)\end{tabular} \\
\midrule
LTF~\citep{py123dgarage} & public checkpoint, no change & 103 & 11 & 0.936 & 0.79 \\
\base & hold, release, lead stop, road guard & 90 & 10 & 0.924 & 0.91 \\
\stD & preliminary decoder, gated & 57 & 7 & 0.933 & 1.45 \\
\stL & release and launch, braking cap, pace gain & 58 & 7 & 0.934 & 1.43 \\
\stC & free-space creep, launch for rolling starts & 58 & 7 & 0.953 & 1.44 \\
\sysname & final decoder & \textbf{45} & 7 & \textbf{0.957} & \textbf{1.86} \\
\bottomrule
\end{tabular}
\end{table}
\begin{table}[!htb]
\centering
\caption{Results on the official test set (nuPlan track, 1,000 hidden scenes), from the final leaderboard of the challenge. Capability score: the ranking score of the leaderboard, fitted jointly over all entries. Distance: the leaderboard's average distance between at-fault incidents. The entries of the organisers are evaluated on the same leaderboard. The public checkpoint is submitted without any change; its capability score is the median over five such entries.}
\label{tab:board}
\small
\begin{tabular}{@{}lrrr@{}}
\toprule
Configuration & \begin{tabular}[b]{@{}r@{}}Capability\\score\end{tabular} & \begin{tabular}[b]{@{}r@{}}Mean scene\\score\end{tabular} & \begin{tabular}[b]{@{}r@{}}Distance per\\incident (km)\end{tabular} \\
\midrule
\multicolumn{4}{@{}l}{\emph{Entries of the organisers}} \\
DiffusionDrive~\citep{liao2025diffusiondrive}, SimScale release~\citep{tian2026simscale} & 1147 & 0.7777 & 0.27 \\
GTRS-Dense~\citep{li2025gtrs}, SimScale release~\citep{tian2026simscale} & 1080 & 0.7624 & 0.41 \\
LTF, best entry of the organisers~\citep{py123dgarage} & 2478 & 0.9437 & 0.98 \\
\midrule
\multicolumn{4}{@{}l}{\emph{Public checkpoint and our entries}} \\
LTF, public checkpoint~\citep{py123dgarage} & 2444 & 0.9205 & 0.83 \\
\early & \textbf{2573} & 0.9693 & 1.77 \\
\base & 2548 & 0.9711 & 1.96 \\
\stL & 2552 & 0.9751 & \textbf{3.55} \\
\sysname & 2535 & \textbf{0.9763} & 2.28 \\
\bottomrule
\end{tabular}
\end{table}

\begin{figure}[t]
\centering
\includegraphics[width=\linewidth]{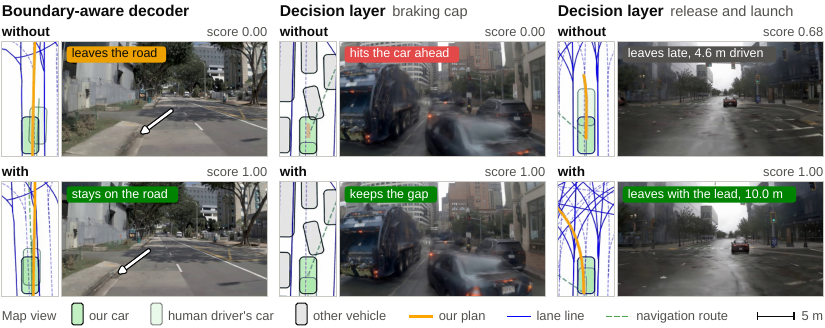}
\caption{Closed-loop examples. Each column shows one scene at the same time step without (top) and with (bottom) the named component. The \decshort keeps the car on the road; the braking cap of the decision layer keeps the gap to a slowing lead vehicle, and the release lets a held car depart with its lead vehicle. In the map views the outline is the human driver's position at that step.}
\label{fig:qual}
\end{figure}

\rhead{Validation set.}
\sysname reduces the incident scenes of the LTF planner from 103 to 45 and raises the mean scene score from 0.936 to 0.957 (Table~\ref{tab:main}).
The paired difference of the mean score is $+0.0207$ with a 95\,\% interval of $[+0.0101, +0.0281]$, and the distance driven per incident scene rises from 0.79\,km to 1.86\,km.
With the basic safeguards alone, \base has fewer incident scenes than LTF but less progress, since a held car falls behind a human driver who departs: the mean score falls from 0.936 to 0.924.
Relative to \base, the gain of \sysname is $+0.0329$ ($[+0.0185, +0.0477]$).
The two components act on different quantities.
The decoder removes incidents: the preliminary decoder takes the incident scenes from 90 to 57 and the final decoder to 45, and its effect is confined to the scenes in which it plans (Appendix~\ref{app:decoder}).
The decision layer mainly recovers progress: the steps of \stC add 0.019 to the mean score without adding an incident scene. Table~\ref{tab:modules} evaluates each of its modules on the scenes of the kind it acts on; none of them adds an incident scene there (Appendices~\ref{app:standstillsec} and~\ref{app:pace}).
Fig.~\ref{fig:qual} shows examples of both components.

\begin{table}[!htb]
\centering
\caption{Modules of the decision layer, each evaluated on scenes of the kind it acts on: mean scene score without and with the module. Gain and lose count the scenes whose score rises or falls. Held start: the car stands at the end of the hand-over. Rolling start: the car moves below 3\ms.}
\label{tab:modules}
\small
\begin{tabular}{@{}llrrrr@{}}
\toprule
Module & Scenes & Without & With & Gain & Lose \\
\midrule
Release and launch & 67 held starts behind a departing lead & 0.654 & 0.731 & 34 & 0 \\
Free-space creep & 541 held starts & 0.539 & 0.686 & 232 & 0 \\
Creep behind a far lead & 541 held starts & 0.686 & 0.717 & 50 & 0 \\
Launch for rolling starts & 1,555 rolling starts & 0.948 & 0.968 & 68 & 1 \\
\bottomrule
\end{tabular}
\end{table}

\rhead{Test set.}
Table~\ref{tab:board} gives the results of the final leaderboard for three entries of the organisers, the public LTF checkpoint and four variants of \sysname.
The organisers' entries with other planners, DiffusionDrive and GTRS-Dense, stay below 0.78 in the mean scene score, and their best LTF entry reaches 0.9437.
With the hold, the release and the road guard alone, \early raises the mean scene score of the planner from 0.9205 to 0.9693 and more than doubles its distance per incident.
\sysname has the highest mean scene score, 0.9763, and drives 2.28\,km per incident, against 0.83\,km for LTF; the longest distance per incident, 3.55\,km, is reached by \stL.
The challenge ranks the teams by a capability score that is fitted jointly over all entries, on scene scores averaged over three runs.
Our team took first place in the nuPlan track~\citep{alpasimchallenge} with \early, whose capability score of 2573 is the highest of the track.
The scores of the four variants, 2535 to 2573, lie within one another's 95\,\% intervals, which are about 300 points wide.

\section{Resources and training}
\label{sec:resources}
\rhead{Data.}
The decoder is trained on 271,408 frames of 1,038 nuPlan training logs~\citep{caesar2021nuplan}, with the mapped drivable area as the label, and on 11,114 closed-loop rollouts of our own system; the preliminary decoder uses the frames only.
The near-field detector is trained on 123,811 frames rendered in our rollouts of the challenge's scene packages.

\rhead{Pretrained weights.}
We start from the public LTF checkpoint \texttt{ltf-base-joint64-68M}~\citep{py123dgarage} and, for the detector, from a SegFormer-B0 fine-tuned on Cityscapes~\citep{xie2021segformer,cordts2016cityscapes}.
All external data and weights are public; Table~\ref{tab:resources} lists them with their licences.

\rhead{Training cost.}
Training is cheap: the decoder takes 4,000 AdamW steps, under half a GPU-hour on one H100 GPU, and the detector about five GPU-minutes (Table~\ref{tab:params}).
The larger cost is the data of the final decoder: collecting its closed-loop rollouts takes about 80 GPU-hours of simulation.

\begin{table}[!htb]
\centering
\caption{Training settings of the \decshort. \sysname uses the final decoder; the preliminary decoder appears only in the ablation. Both are trained with AdamW, a learning rate of $10^{-5}$ and a batch size of 64.}
\label{tab:params}
\footnotesize
\begin{tabular}{@{}L{3.7cm}L{5.2cm}L{4.1cm}@{}}
\toprule
 & Final decoder & Preliminary decoder \\
\midrule
Initial weights & public LTF decoder & public LTF decoder \\
Training data & 271,408 frames of 1,038 nuPlan logs and 11,114 closed-loop rollouts & the same frames \\
Distillation weights $\lambda_{\mathrm{s}}$, $\lambda_{\ell}$ & 90, 18 & 90, 18 \\
Boundary hinge $\lambda_{\mathrm{h}}$ (margin) & 30 (0.15\,m) & 100 (0.15\,m) \\
Road-user hinge $\lambda_{\mathrm{a}}$ (margin) & 30 (0.5\,m) & -- \\
Time weights $w_k$ & \multicolumn{2}{l}{1 up to 3\,s, 0.5 from 3 to 4\,s} \\
Training steps & 4,000 & 4,000 \\
Weights used & after 2,000 steps, interpolated with the public decoder (factor 0.75) & after 4,000 steps \\
GPU time & under 0.5\,h of training; about 80\,h of simulation for the rollouts & under 0.5\,h of training \\
\bottomrule
\end{tabular}
\end{table}

\section{Runtime and reproducibility}
\label{sec:runtime}
\rhead{Serving.}
The decoder gate decides before the decoders run, so one decoder runs on most replans; the detector runs only while a car is held, in 4\,\% of the replans.

\rhead{Latency and memory.}
On an H100 GPU the 95th-percentile time per replan is typically 50 to 64\,ms, and 86 to 88\,ms in held starts, where the detector runs, against the challenge's target of 100\,ms. GPU memory peaks at 1.8\,GiB with both decoders, below the 16\,GiB limit.

\rhead{Reproducibility.}
Our closed-loop results are repeatable: a repeated run of 1,198 scenes reproduced every per-scene result. Our code is available at \url{https://github.com/JiahaoShu/GuardLTF}; its configuration file lists every parameter of the decision layer with its unit.

\section{Conclusion, limitations and safety considerations}
\label{sec:conclusion}
We analysed the closed-loop failures of a public camera-only planner and found two patterns: shallow off-road exits at road edges that the planner's own map does not show, and stopping and starting that do not follow the human driver's behaviour around a lead vehicle.
\sysname addresses them with a \decshort that trains the margin to the road edge into the plan and a \dlname around the frozen planner.
On a validation set of 3,625 scenes it reduces the incident scenes of the planner from 103 to 45 and raises the mean scene score from 0.936 to 0.957, and on the official test set it raises the mean scene score from 0.9205 to 0.9763 and the distance per incident from 0.83\,km to 2.28\,km.
Neither component modifies the planner; both need only a trajectory decoder and a BEV map.

\rhead{Limitations.}
The decision layer is a fixed set of hand-designed conditions: each encodes one situation that we observed, and their thresholds are tuned for this simulator and replan period, so the layer does not adapt to situations outside this set.
Learning these decisions is the natural next step. A decision module trained on closed-loop outcomes could choose between the plans, stop the car and release it from the scene itself, and so adapt to new scenes without new rules; the rollouts of the present system provide training data for it, and the hand-designed conditions a strong baseline and explicit safety constraints.
Off-road exits remain the dominant failure: they account for 38 of the 45 incident scenes of \sysname on the validation set, and the most common remaining collision is a contact with a stationary vehicle at the edge of the planned path.
The gain of the final decoder over the preliminary one is not uniform across scenes (Appendix~\ref{app:decoder}), and on the test set \stL reaches a longer distance per incident than \sysname.
All results are for one planner and one simulator with scenes of 5\,s.

\rhead{Safety considerations.}
Holding a standing car is conservative towards traffic ahead but can expose it to traffic from behind; the free-space creep halves the number of scenes in which the replayed following vehicle runs into the held car (Appendix~\ref{app:standstillsec}).
Cyclists are not represented in the BEV map that the decision layer reads.
If a module of the decision layer raises an internal error, the plan is left unchanged.
The system has been evaluated in simulation only, and the fixed conditions of the decision layer do not replace a safety case.

\appendix
\section{Scene sets and protocol}
\label{app:sets}
\rhead{Scenes.}
All local results are closed-loop runs in the official simulator, controller and scorer~\citep{alpasimcode} in the challenge's nuPlan configuration, with one rollout per scene.
We use 11,146 scenes from the official scene packages and divide them into 7,521 \emph{development scenes} and a \emph{validation set} of 3,625 scenes from 21 recording sessions.
The two parts share no scene.
The preliminary decoder is trained on nuPlan training logs only, and the final decoder adds rollouts of development scenes.
The gates of the preliminary decoder were set on the development scenes alone; the other components were selected with access to the validation set, which therefore serves for model selection and ablation.
The \emph{test set} is the hidden set of 1,000 scenes behind the official leaderboard of the challenge, which is evaluated by the organisers.

\rhead{Scoring.}
The score of a scene is zero after an at-fault collision, an off-road exit or an exit from the corridor of 4\,m to either side of the human path, and $\min(\text{progress}/0.8,\,1)$ otherwise, where progress is measured against the human driver.
We call a scene with an at-fault collision or an off-road exit an \emph{incident scene}; corridor exits are rare and are reported separately.

\rhead{Comparisons.}
We report the number of incident scenes, the at-fault collisions among them, the mean scene score and the distance driven per incident scene.
Comparisons are paired, that is, two configurations are compared on the same scenes; the simulation is repeatable, and a repeated run of 1,198 scenes reproduced every per-scene result.
Scenes of one recording session are not independent, so confidence intervals are bootstrap intervals over whole recording sessions (10,000 resamples).

\rhead{Configurations.}
Table~\ref{tab:main} starts from the public LTF checkpoint without any change and from \base, and adds the components in four cumulative stages.
\stD adds the preliminary decoder with its gates.
\stL adds the release and launch, the braking cap and the pace gain.
\stC adds the free-space creep and the launch for rolling starts.
\sysname replaces it by the final decoder.

\section{Diagnosis of the LTF planner}
\label{app:diagnosis}
The LTF planner fails in two ways in closed loop, and each component of \sysname answers one of them.

\rhead{Road edge.}
Off-road exits are the most frequent failure: on 1,198 development scenes, 23 of the 29 zero scores of \base are exits in which the planned path itself leads the car over the road edge.
The exits are shallow. Of 69 off-road exits that remain with the preliminary decoder, 11 occur on the human driver's own path at the first off-road step and 32 more within 0.7\,m of it.
The planner's BEV map mostly does not show the edge that the plan crosses: a guard that moves plans away from cells predicted as off-road acts in only 27 of 113 exits of \base.
A check on the predicted map therefore repairs a minority of the exits, which motivates training the margin to the road edge into the plan itself (Appendix~\ref{app:decoder}).

\rhead{Stopping behind a vehicle.}
Half of the at-fault collisions that remain once the road-edge errors are reduced are front collisions in which the car drives past the point where the human driver stops behind a vehicle.
In 5 of 12 such collisions the hand-over, which replays the human driver, already shows the braking; the braking cap uses this cue (Appendix~\ref{app:pace}).

\rhead{Starting from rest.}
At a standstill, the two input frames carry little evidence of when to depart.
The planner waits behind a stopped lead vehicle that its map shows, and it rarely misses one (in 8 of 261 starts at rest with a vehicle within 10\,m in the recording).
Where no lead is visible, however, its plan departs before the human driver: on 443 scenes in Las Vegas the plan accelerates before the human driver departs in 180 of the 201 starts at rest, while with a vehicle within 10\,m the planner never leaves before the human driver.
Departing early is costly, because the corridor follows the human trajectory.
The recorded drives give no simple cue for the departure either: in 25,057 standing moments of the public recordings the human driver stays for the next 5.5\,s in 79--94\,\% of the cases at every gap to the vehicle ahead, and still in about half of the 1,426 moments with neither a red light on the route nor a vehicle within 15\,m.
We therefore do not try to predict departures. The decision layer instead bounds the cost of a wrong decision: it holds a standing car, releases it when the lead vehicle departs, and otherwise allows only a short creep on a road that is confirmed to be free (Appendix~\ref{app:standstillsec}).
Holding alone costs progress: on the validation set \base has 90 incident scenes against 103 for the planner alone, but a lower mean score, 0.924 against 0.936 (Table~\ref{tab:main}), because a held car falls behind when the human driver departs.
The release and the creep are designed to recover this progress.

\section{\Decname{}}
\label{app:decoder}
The planner's map does not show the road edge that its plans cross (Appendix~\ref{app:diagnosis}), so the margin is trained into the plan: the drivable-area boundary of the nuPlan maps is a training label for a second decoder and is not available to the policy at run time.

\rhead{Training.}
The loss is given in Eq.~\ref{eq:loss} (Section~\ref{sec:policy}); we refer to its hinge term as the \declossname.
Table~\ref{tab:params} lists the training settings.
Like the preliminary decoder, the final decoder starts from the public weights; it adds 11,114 closed-loop rollouts, 171 of them with an off-road exit, and the hinge to other road users, with a weight of 30 for both hinges.
The rollouts were collected in two iterations: 6,333 with \base and 4,781 with a decoder trained on the first set.
We use the weights after 2,000 steps, interpolated with the weights of the public decoder with a factor of 0.75~\citep{wortsman2022robust}.

\rhead{Weight of the hinge.}
For a decoder trained on recorded drives only, on nine simulated off-road exits from log files excluded from training, hinge weights of 0, 1 and 10 remove none, a weight of 30 removes six, and weights of 60 and 100 remove eight.
The drift of plans that need no correction grows with the weight, from 0.04\,m to 0.21\,m at the 95th percentile.
The preliminary decoder uses $\lambda_{\mathrm{h}} = 100$, which performed best in closed loop on the development scenes; with rollouts in the training data a weight of 30 suffices, and the final decoder uses it.

\begin{figure}[!htb]
\centering
\includegraphics[width=\linewidth]{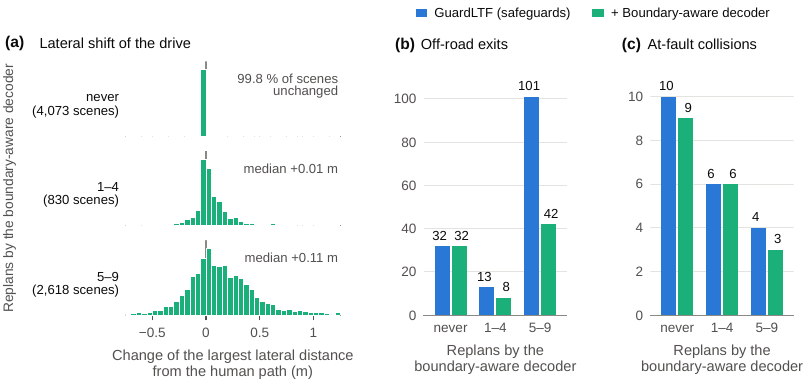}
\caption{Effect of the preliminary \decshort on the 7,521 development scenes, grouped by how many of the ten replans of a scene the decoder plans. (a) Change of the largest lateral distance from the human path. (b) Off-road exits. (c) At-fault collisions.}
\label{fig:kmeffect}
\end{figure}

\begin{figure}[!htb]
\centering
\includegraphics[width=\linewidth]{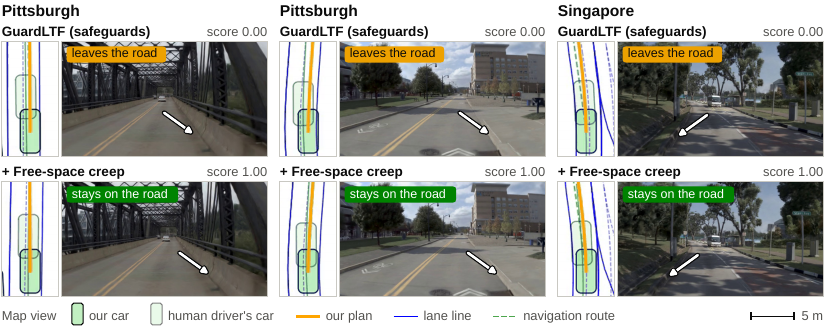}
\caption{Three off-road exits of \base that the preliminary decoder removes, at 4.5\,s. Top: \base; bottom: \cfgcrp, a configuration with the decoder. Arrows mark the same road edge in both frames. The decoder plans 5, 9 and 9 of the ten replans (left to right).}
\label{fig:qdecoder}
\end{figure}

\rhead{Where the decoder acts.}
Under its gates the preliminary decoder plans about one third of the replans.
Fig.~\ref{fig:kmeffect} groups the development scenes by the number of replans that the decoder plans.
In the 4,073 scenes in which it never plans, 99.8\,\% of the results are unchanged.
In the 2,618 scenes in which it plans at least five of the ten replans, the off-road exits fall from 101 to 42, with a median lateral shift of 0.11\,m.
A shift of about a decimetre thus removes more than half of the exits, in line with the diagnosis of Appendix~\ref{app:diagnosis}.
Over all development scenes, off-road exits fall from 146 to 82 and at-fault collisions from 20 to 18; Fig.~\ref{fig:qdecoder} shows three of the exits that the decoder removes.

\rhead{Gates.}
The gates limit the collisions that a lateral shift can cause.
The preliminary decoder, whose loss has no term for other road users, has a third gate besides the two of Section~\ref{sec:policy}: a lead vehicle at the start of a scene hands the whole scene to the nominal plan (\emph{lead gate}).
Without any gate, an earlier decoder trained with a hinge weight of 30 raises the at-fault collisions on 1,198 development scenes from 5 to 9.
On 4,188 development scenes the preliminary decoder has 19 at-fault collisions with the proximity gate alone and 11 with all gates, against 10 for \base.

\rhead{Final decoder.}
On the validation set, training on closed-loop rollouts with the road-user hinge reduces the incident scenes from 58 to 48, and removing the lead gate, which this hinge makes possible, gives 45, with 7 at-fault collisions as before; the decoder then plans 44\,\% of the replans instead of 39\,\%.
Without the road-user hinge it reaches 52 instead of 48 incident scenes.
The paired gain of \sysname over \stC in the mean score is $+0.0041$ ($[+0.0017, +0.0067]$).
The gain is not uniform: on 1,214 development scenes whose log files are excluded from the training of both decoders, the final decoder has 16 incident scenes against 12, with one at-fault collision fewer.

\rhead{Targeted checks.}
\stD also contains six small refinements of the gates and of the stop behind a lead vehicle, each derived from individual incidents.
They remove 10 incident scenes on the scenes they were derived from and none of the 25 incident scenes of 1,791 further scenes, so they add little to the gain of the decoder and its gates.

\section{Standstill module}
\label{app:standstillsec}\label{app:creep}
The standstill module decides when a standing car may move (Section~\ref{sec:policy}); Appendix~\ref{app:diagnosis} shows why the planner cannot.
A held car is released when the planner's own plan accelerates and the lead vehicle has moved away; the free-space creep applies where the BEV map shows no lead, or only a far one.
The subsections below discuss the modules of Table~\ref{tab:modules}.

\subsection{Release and launch}
\begin{figure}[!htb]
\centering
\includegraphics[width=\linewidth]{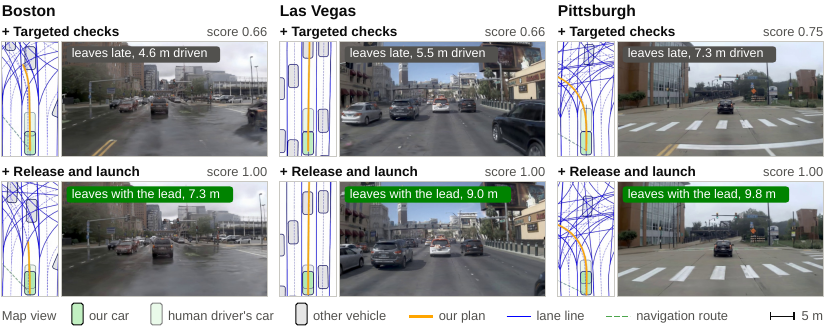}
\caption{Release and launch in three scenes, at 5.0\,s. Top: with the release of \base the car departs late and has driven 4.6 to 7.3\,m. Bottom: released at a recession of the lead of 0.5\,m, it has driven 7.3 to 9.8\,m and is closer to the human driver's position (outline in the map view).}
\label{fig:qrelease}
\end{figure}

With the release of \base, a held car departs late or not at all behind a departing lead.
Releasing once the lead has receded by 0.5\,m and then following it raises the mean score of 67 such scenes from 0.654 to 0.731; half of the scenes gain and none loses (Fig.~\ref{fig:qrelease}).
The bumper gap after a release stays at or above 3.1\,m.
A threshold of 0.2\,m scores higher on these scenes (0.762); we use the more conservative 0.5\,m.

\subsection{Free-space creep: moving off on a free road}
\begin{figure}[!htb]
\centering
\includegraphics[width=\linewidth]{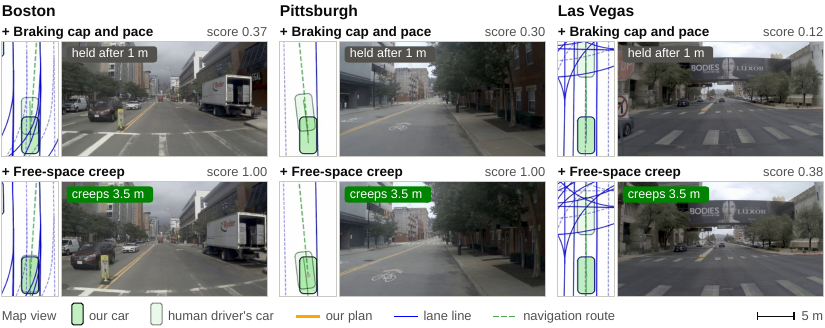}
\caption{Free-space creep in three held starts without a lead in the BEV map, at 5.0\,s. Where the human driver covers 5.8 or 5.9\,m, the creep of 3.5\,m earns the full score; where the human driver covers 14.6\,m, it earns 0.38.}
\label{fig:qcreep}
\end{figure}

Held starts without a lead in the BEV map are the largest remaining loss of progress, since the car is held for the whole scene while the human driver departs.
The creep raises the mean score of the 541 held starts from 0.539 to 0.686 (Fig.~\ref{fig:qcreep}).
The detector grants the creep in 239 of the 244 held starts without a lead, and offline it reports a free road in 3 of 1,334 test frames with a vehicle within 5\,m (0.22\,\%, Table~\ref{tab:detector}).
On the 132 held starts that lie on the detector's test log files, the mean score rises from 0.690 to 0.780.
Extending the creep to starts with a lead at 12.5\,m or more, together with a longer window for the release, adds a further 0.031.
The creep also halves the number of scenes in which the replayed following vehicle runs into the held car, from 76 to 39.

\rhead{Near-field detector.}
Table~\ref{tab:detector} summarises the detector of Section~\ref{sec:policy}.
Its data are split by log file, and the operating point was fixed before the first closed-loop run.

\begin{table}[!htb]
\centering
\caption{The near-field detector. Frames are rendered in our closed-loop rollouts and labelled with the actor boxes of the simulator.}
\label{tab:detector}
\footnotesize
\begin{tabular}{@{}L{5.8cm}L{7.4cm}@{}}
\toprule
Quantity & Value \\
\midrule
Network & SegFormer MiT-B0 encoder~\citep{xie2021segformer} with a head of eight ordered outputs: gap ahead of the bumper of at least 2, 3, 4, 5, 6, 8, 10 or 15\,m \\
Input & crop of the front-camera frame, 448 by 256 pixels \\
Data & 123,811 frames of 11,261 rollouts on 136 log files; split by log file into 49,250 training, 41,022 validation and 33,539 test frames \\
Operating point & the road counts as free when the logit of ``gap of at least 8\,m'' exceeds 3.5 \\
\midrule
Standstill test frames with a vehicle within 5\,m, reported free & 3 of 1,334 (0.22\,\%) \\
Standstill test frames without a vehicle within 10\,m, reported free & 82\,\% of 863 \\
Area under the curve of the output on standstill test frames & 0.997 \\
\bottomrule
\end{tabular}
\end{table}

\section{Moving-safety module: braking cap and pace}
\label{app:pace}\label{app:braking}
The moving-safety module checks the plan that the decoder gate selects (Section~\ref{sec:policy}); this appendix gives the results of its three re-timing steps.

\begin{figure}[!htb]
\centering
\includegraphics[width=\linewidth]{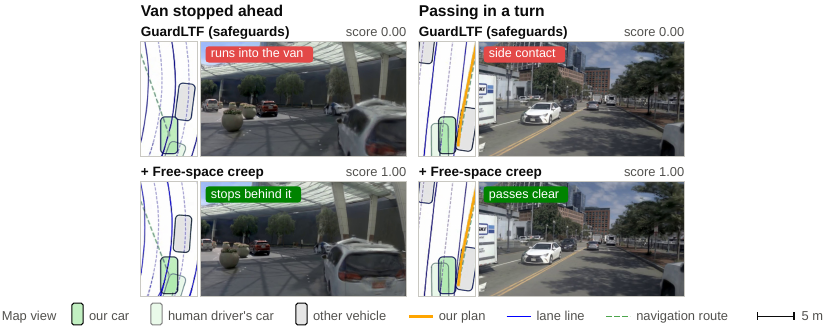}
\caption{Two further at-fault collisions of \base that \stC avoids (the braking cap is shown in Fig.~\ref{fig:qual}). Left: the car stops behind a stopped van. Right: a side contact with the vehicle on the right while passing in a turn; this contact returns with the final decoder.}
\label{fig:qcollisions}
\end{figure}

\rhead{Braking cap and pace gain.}
The braking cap targets the front collisions of Appendix~\ref{app:diagnosis}: it limits the travel distance when the hand-over shows hard braking behind a slow lead.
Over the local scenes it removes three at-fault collisions and costs the full score in one scene (Fig.~\ref{fig:qcollisions}).
The pace gain of 3\,\% applies on a free, straight road; it brings eight development scenes to the full score and adds one off-road exit on the validation set.

\rhead{Launch for rolling starts.}
In scenes that start below 3\ms with the human driver accelerating, the planner often crawls or stops.
The launch raises the mean score of 1,555 such scenes from 0.948 to 0.968 (Table~\ref{tab:modules}).
It is restricted to straight plans without a vehicle within 15\,m; without this restriction the mean score is 0.971 at the cost of one additional incident scene.

\section{Closed-loop results in full}
\label{app:results}
\begin{figure}[!htb]
\centering
\includegraphics[width=\linewidth]{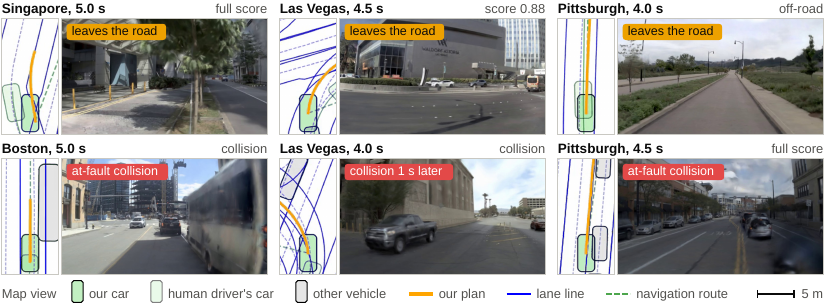}
\caption{Incidents of \stC: off-road exits (top) and at-fault collisions (bottom), each at the step of the event. The label on the right gives the outcome of \base in the same scene; in three of the six scenes \base has no incident. \sysname no longer has the Las Vegas exit and the Pittsburgh collision.}
\label{fig:qfail}
\end{figure}

\begin{table}[!htb]
\centering
\caption{Results on the validation set (3,625 scenes) for every step. The steps below \base are cumulative; the qualitative figures use the same labels. The last column gives the stage of Table~\ref{tab:main} that ends with the step. Distance: km driven per incident scene.}
\label{tab:steps}
\small
\begin{tabular}{@{}lrrrrl@{}}
\toprule
Step & \begin{tabular}[b]{@{}r@{}}Incident\\scenes\end{tabular} & \begin{tabular}[b]{@{}r@{}}At-fault\\collisions\end{tabular} & \begin{tabular}[b]{@{}r@{}}Mean\\score\end{tabular} & \begin{tabular}[b]{@{}r@{}}Distance\\(km)\end{tabular} & Stage \\
\midrule
LTF, public checkpoint & 103 & 11 & 0.936 & 0.79 & LTF \\
\base & 90 & 10 & 0.924 & 0.91 & \base \\
\cfgtgt & 57 & 7 & 0.933 & 1.45 & \stD \\
\cfgrel & 57 & 7 & 0.934 & 1.45 & \\
\cfgbrk & 58 & 7 & 0.934 & 1.43 & \stL \\
\cfgcrp & 58 & 7 & 0.945 & 1.43 & \\
\cfgrol & 58 & 7 & 0.950 & 1.44 & \\
\cfgfar & 58 & 7 & 0.953 & 1.44 & \stC \\
+\,Final decoder & 48 & 8 & 0.955 & 1.74 & \\
+\,Lead gate off & 45 & 7 & 0.957 & 1.86 & \sysname \\
\bottomrule
\end{tabular}
\end{table}
Table~\ref{tab:steps} lists the individual steps behind the stages of Table~\ref{tab:main}.
\cfgtgt is the preliminary decoder with its gates and the six targeted checks.
The steps of the standstill module and the launch for rolling starts change only scenes that start at rest or at low speed, and the rows of these steps are obtained by rerunning the scenes that the step can change.
The rows of LTF, \base and \sysname are full runs of the validation set.
For \sysname, the full run agrees with the result obtained by rerunning in 3,619 of the 3,625 scenes; in the other six scenes the progress differs and no incident is added or removed.
From LTF to \sysname the score rises in 411 scenes and falls in 176.
Fig.~\ref{fig:qfail} shows incidents that remain.

\section{Public resources}
\label{app:runtime}\label{app:resources}
Table~\ref{tab:resources} lists the public resources used. All external data and weights are public; the final decoder and the detector additionally use our own closed-loop rollouts of the public scenes.

\begin{table}[!htb]
\centering
\caption{Public resources and their licences. HF: Hugging Face.}
\label{tab:resources}
\footnotesize
\begin{tabular}{@{}L{3.1cm}L{3.2cm}L{6.6cm}@{}}
\toprule
Resource & Use & Source; licence \\
\midrule
nuPlan dataset & training data; scenes & HF \texttt{kesai-labs/nuplan}~\citep{caesar2021nuplan}; CC BY-NC-SA 4.0 and nuPlan Terms of Use \\
NAVSIM navtest split & scene lists of part of the development scenes & \citep{dauner2024navsim}; Apache-2.0 \\
AlpaSim nuPlan scenes & local scenes; frames and labels of the detector & HF \texttt{OpenDriveLab/\allowbreak AlpasimChallenge2026\_\allowbreak nuplan\_track}; CC BY-NC-SA 4.0 \\
py123d-garage & policy code; LTF checkpoint & \citep{py123dgarage}; Apache-2.0 (code), CC BY-NC-SA 4.0 (checkpoint) \\
SegFormer-B0, Cityscapes & initial weights of the detector's encoder & HF \texttt{nvidia/\allowbreak segformer-b0-\allowbreak finetuned-\allowbreak cityscapes-\allowbreak 1024-1024}~\citep{xie2021segformer,cordts2016cityscapes}; NVIDIA Source Code License (non-commercial) \\
AlpaSim & simulator, controller, scorer & \citep{alpasimcode}; Apache-2.0 \\
\bottomrule
\end{tabular}
\end{table}

\end{document}